\documentclass[letterpaper]{article} % DO NOT CHANGE THIS
\usepackage{aaai2027}  % DO NOT CHANGE THIS
\usepackage[hyphens]{url}  % DO NOT CHANGE THIS
\usepackage{graphicx} % DO NOT CHANGE THIS
\usepackage{natbib}  % DO NOT CHANGE THIS AND DO NOT ADD ANY OPTIONS TO IT
\usepackage{caption} % DO NOT CHANGE THIS AND DO NOT ADD ANY OPTIONS TO IT
\usepackage{algorithm}
\usepackage{algorithmic}
\usepackage{booktabs}
\usepackage{multirow}
\usepackage{amsmath,amsfonts,amssymb}

\newcommand{\kopemem}{KOPE-Mem}

\title{Beyond Scaling: Self-Evolving LLM Agents for Hardware Kernel Optimization via an Experience-Driven Workflow and Experience Graph Memory}

\makeatletter
\author{%
    \global\aaai@corrmultitrue%
    Siyuan~Chen\textsuperscript{\rm 1}\equalcontrib,
    Runlin~Hou\textsuperscript{\rm 1}\equalcontrib,
    Shenxiu~Wu\textsuperscript{\rm 1},
    Yansong~Sun\textsuperscript{\rm 1},
    Junming~Cao\textsuperscript{\rm 2},
    Yiyu~Zhang\textsuperscript{\rm 2},
    Shudi~Shao\textsuperscript{\rm 2},
    Junhao~Qiu\textsuperscript{\rm 1},
    Zhichao~Lu\textsuperscript{\rm 1}\corresponding,
    Qingfu~Zhang\textsuperscript{\rm 1}\corresponding
}
\makeatother
\affiliations{
    \textsuperscript{\rm 1}Department of Computer Science, City University of Hong Kong\\
    \textsuperscript{\rm 2}Application Software Engineering Lab, Huawei Technologies Ltd.\\
    siyuan.chen@my.cityu.edu.hk, zhichao.lu@cityu.edu.hk, qingfu.zhang@cityu.edu.hk
}

\begin{document}

\nocopyright
\maketitle

\begin{abstract}
Hardware kernel optimization requires repeated compilation, correctness testing, profiling, and revision. LLM agents can automate parts of this process, and stronger foundation models, longer context windows, and longer execution horizons have improved optimization within individual tasks.
These advances alone do not enable an agent to learn from completed optimization runs. Existing kernel-optimization agents seldom preserve a decision, its observed execution feedback, and the later decisions that use that evidence. Retaining every prior trajectory is also impractical because an expanding history competes with the current task for context.
We present KOPE, an experience-driven framework for hardware kernel optimization. KOPE records optimization trajectories with correctness and performance feedback in Experience Graph Memory, then uses Active Context Management and Injection to retrieve relevant experience under a fixed token budget. The graph retains decision order, observed outcomes, and alternative branches, allowing evidence collected on the target hardware to inform later optimization steps and tasks.
Under the same GLM-5.2 setting, the geometric mean of KOPE's per-operator speedups is $1.54\times$ that of CANNBot, the strongest competing baseline. In a complete 53-operator ablation, Active Context Management and Injection raises pass rate from 60.0\% to 84.6\%, increases the evaluator-reported positive-field geometric mean from 0.0382 to 0.0661, and reduces optimization token consumption from 15.9B to 1.113B tokens relative to passive agent-led context construction. Enabling Experience Graph Memory raises full-suite pass rate from 55.2\% to 84.6\% and yields a $1.43\times$ geometric-mean speedup on valid timing comparisons. These results support continual optimization through external experience while the foundation model remains fixed.
\end{abstract}

% =============================================================================
\section{Introduction}
\label{sec:introduction}
% =============================================================================

Hardware kernel optimization remains an expertise-intensive process. Performance depends on the memory hierarchy, data movement, parallel execution, and constraints specific to the target device. A candidate implementation must be compiled, tested for correctness, benchmarked, and often revised several times. Recent LLM agents automate parts of this process through stronger foundation models, domain adaptation, retrieval, larger context windows, and longer execution trajectories~\cite{ouyang2025kernelbench,wei2025astra,ascendkernelgen2026}.

Their effectiveness still depends heavily on hardware knowledge encoded in the underlying model. That dependence is particularly restrictive for newly introduced hardware with few public implementations or optimization traces. Automated optimization is useful precisely during this early stage, before the software ecosystem has accumulated a substantial public corpus. The gains reported for chip-domain and hardware-specific adaptation show that target-specific knowledge matters~\cite{liu2023chipnemo,ascendkernelgen2026}.

Recent systems use domain adaptation, agentic reinforcement learning, multi-agent planning, and execution-guided iteration to strengthen model capability or refine a kernel within a task~\cite{wei2025astra,tehrani2026makora,ascendkernelgen2026,dai2026cudaagent}. Their execution feedback mainly guides the current run and is rarely organized as reusable evidence for later decisions or tasks. The agent therefore remains dependent on knowledge already present in the model. Retaining every trajectory does not solve this problem: an expanding history consumes context and may obscure evidence relevant to the current task~\cite{liu2024lostmiddle,hsieh2024ruler}. This leads to the following research question:

\emph{How can an LLM agent turn execution feedback from exploration into reusable knowledge and apply it across optimization steps and tasks on hardware with scarce public training data?}

KOPE addresses this question through an agent workflow that learns during exploration. It converts compiler diagnostics, correctness outcomes, profiling observations, and measured performance changes into external knowledge while keeping the foundation model fixed. \emph{Experience Graph Memory} preserves decision-to-outcome histories and alternative branches. \emph{Active Context Management and Injection} selects the experience relevant to the current optimization state under a bounded token budget. Knowledge collected on the target hardware then informs later decisions and optimization runs. Our contributions comprise one system-level workflow and two supporting mechanisms:

\begin{itemize}

\item \textbf{Experience-Driven Optimization Workflow.}
We develop a closed-loop agent workflow that validates each candidate for correctness and performance. The workflow records the resulting trajectory and reuses that evidence in later decisions and runs.

\item \textbf{Experience Graph Memory.}
We introduce \kopemem{}, which organizes optimization decisions and measured outcomes as a directed experience graph. Provenance links preserve execution order and alternative branches, so retrieval can consider downstream outcomes as well as textual relevance.

\item \textbf{Active Context Management and Injection.}
We select and inject task-relevant knowledge under the available token budget. The assembled prompt retains the required task state and useful prior experience without carrying the full history.

\end{itemize}

We evaluate workflow-level learning on Huawei Ascend NPUs using AscendC kernels~\cite{liao2021ascend}. Public AscendC implementations and optimization traces are limited, while CANN Bench supplies a complete environment for compilation, functional testing, and performance measurement~\cite{gao2026cannbench}. This combination lets us study whether an agent can improve from evidence generated on the target hardware.

With GLM-5.2, KOPE attains a per-operator speedup geometric mean that is $1.54\times$ that of CANNBot. In the complete 53-operator context-policy ablation, Active Context Management and Injection raises pass rate from $60.0\%$ to $84.6\%$, increases the positive-field geometric mean by $1.73\times$, and reduces optimization token consumption from 15.9B to 1.113B tokens relative to passive agent-led context construction. The Experience Graph Memory ablation yields a $1.43\times$ geometric-mean speedup on 412 paired valid timing cases. CUDA-Agent, a strong NVIDIA-GPU optimization workflow, passes $14.7\%$ and $9.1\%$ of the full Ascend suite with GLM-5.2 and Deepseek-V4-Pro, respectively, but neither configuration solves a complete operator. These results suggest that model capability and within-task refinement are insufficient when target-specific public corpora are sparse. Effective workflows in this setting should also retain and reuse knowledge obtained during exploration.\footnote{To assess generalization beyond the primary AscendC hardware and software environment, the appendix reports additional optimization results for Triton kernels on RISC-V hardware.}

The remainder of the paper reviews related work, describes KOPE, presents the experimental evaluation, and discusses the scope and limitations of the evidence.

% =============================================================================
\section{Related Work}
\label{sec:related}
% =============================================================================

\paragraph{LLM-based kernel optimization.}
General code models such as Codex and DeepSeek-Coder established broad program-generation capabilities~\cite{chen2021evaluating,guo2024deepseek}, and ChipNeMo demonstrated the value of domain adaptation for chip-design tasks~\cite{liu2023chipnemo}. KernelBench evaluates both functional correctness and speed for LLM-generated GPU kernels~\cite{ouyang2025kernelbench}. CANN Bench extends execution-based evaluation to Ascend NPUs with 53 operators, 1{,}060 public test cases, and a composite score based on compilation, correctness, and performance~\cite{gao2026cannbench}. Recent systems improve different stages of kernel optimization. Astra coordinates specialized agents through repeated generation, testing, profiling, and planning~\cite{wei2025astra}. Makora uses reinforcement-learning post-training, CuSeT applies CUDA-sensitive instruction tuning, and QiMeng-Kernel separates high-level policy from stepwise implementation~\cite{tehrani2026makora,chen2026cuset,zhu2025qimeng}. AscendKernelGen combines domain-specific data, model adaptation, and execution-based evaluation for Ascend NPUs~\cite{ascendkernelgen2026}. KOPE addresses a different question: how execution feedback from one trajectory can become persistent evidence for later optimization runs.

\paragraph{Retrieval and agent memory.}
RAG combines parametric generation with ranked retrieval from non-parametric memory~\cite{lewis2020rag}, while RepoCoder couples retrieval and generation for repository-level code completion~\cite{zhang2023repocoder}. Hierarchical Context Pruning uses repository topology and function-level relevance to build compact completion prompts~\cite{zhang2025hcp}. Agent memory systems add persistence across interactions. MemGPT manages bounded context through an OS-inspired hierarchy~\cite{packer2023memgpt}. ExpeL retains task trajectories and distills reusable insights separately~\cite{zhao2024expel}. ReMe distills, adapts, and refines procedural memories over time~\cite{cao2026reme}. KOPE uses \kopemem{} as the storage substrate for Experience Graph Memory. Narrative Journals and structured Cases adapt ReMe's procedural-memory mechanisms to kernel optimization and connect records through measured outcomes and provenance links. Active Context Management and Injection uses this retrieval interface to assemble a task-conditioned prompt under a token budget.

\paragraph{Long-context utilization and cache selection.}
Longer context windows do not ensure effective use of all included information. ``Lost in the Middle'' found that performance often declines when relevant evidence appears near the middle of a long context~\cite{liu2024lostmiddle}. RULER likewise showed that usable context can be shorter than the advertised window on retrieval, multi-hop, aggregation, and question-answering tasks~\cite{hsieh2024ruler}. These studies establish a context-utilization problem without attributing it to a single mechanism. Bui et al. identify attention dilution from irrelevant tokens as one source and show that learned KV-cache eviction can match or surpass full-cache inference~\cite{bui2026makeeachcount}. KV-cache eviction selects internal representations during inference. KOPE selects external material before the model call. The mechanisms act at different stages and can be combined.

% =============================================================================
\section{KOPE: Experience-Driven Agent Workflow}
\label{sec:framework}
% =============================================================================

\subsection{Learning from Exploration}

KOPE treats every optimization attempt as a learning event. Execution feedback informs both repair of the current candidate and later decisions. For task $x_t$ at iteration $k$, let $s_{t,k}$ contain the kernel specification, incumbent code, latest execution feedback, and current optimization target, let $\mathcal{M}_{t,k}$ denote persistent external memory, and let $B$ be the context budget. One iteration is
\[
\begin{aligned}
z_{t,k} &= R(s_{t,k},\mathcal{M}_{t,k}), \\
\mathcal{C}_{t,k} &= I(s_{t,k},z_{t,k};B), \\
a_{t,k} &= \pi_{\theta}(\mathcal{C}_{t,k}),\qquad y_{t,k}=E(a_{t,k}), \\
\mathcal{M}_{t,k+1} &= U(\mathcal{M}_{t,k},s_{t,k},a_{t,k},y_{t,k}),
\end{aligned}
\]
where $z_{t,k}$ is retrieved experience, $I$ assembles the prompt, $\pi_{\theta}$ is the fixed-parameter LLM agent, and $E$ returns compiler diagnostics, correctness, and measured performance. The updated memory is available when constructing $\mathcal{C}_{t,k+1}$ and persists after the task, so $\mathcal{M}_{t+1,0}=\mathcal{M}_{t,K_t}$ for a task with $K_t$ iterations. In this paper, \emph{self-evolving} refers to changes in external experience and supplied context. The model parameters remain fixed.

The execution loop applies correctness before performance. KOPE compiles and tests each candidate, replacing the incumbent only when a correct candidate improves the measured objective. It records every attempt: successful transformations become positive evidence, while compilation failures, incorrect outputs, and regressions remain available as failure evidence. The next iteration can respond to an observed failure or extend a successful decision immediately.

\begin{figure*}[t]
\centering
\includegraphics[width=0.98\textwidth]{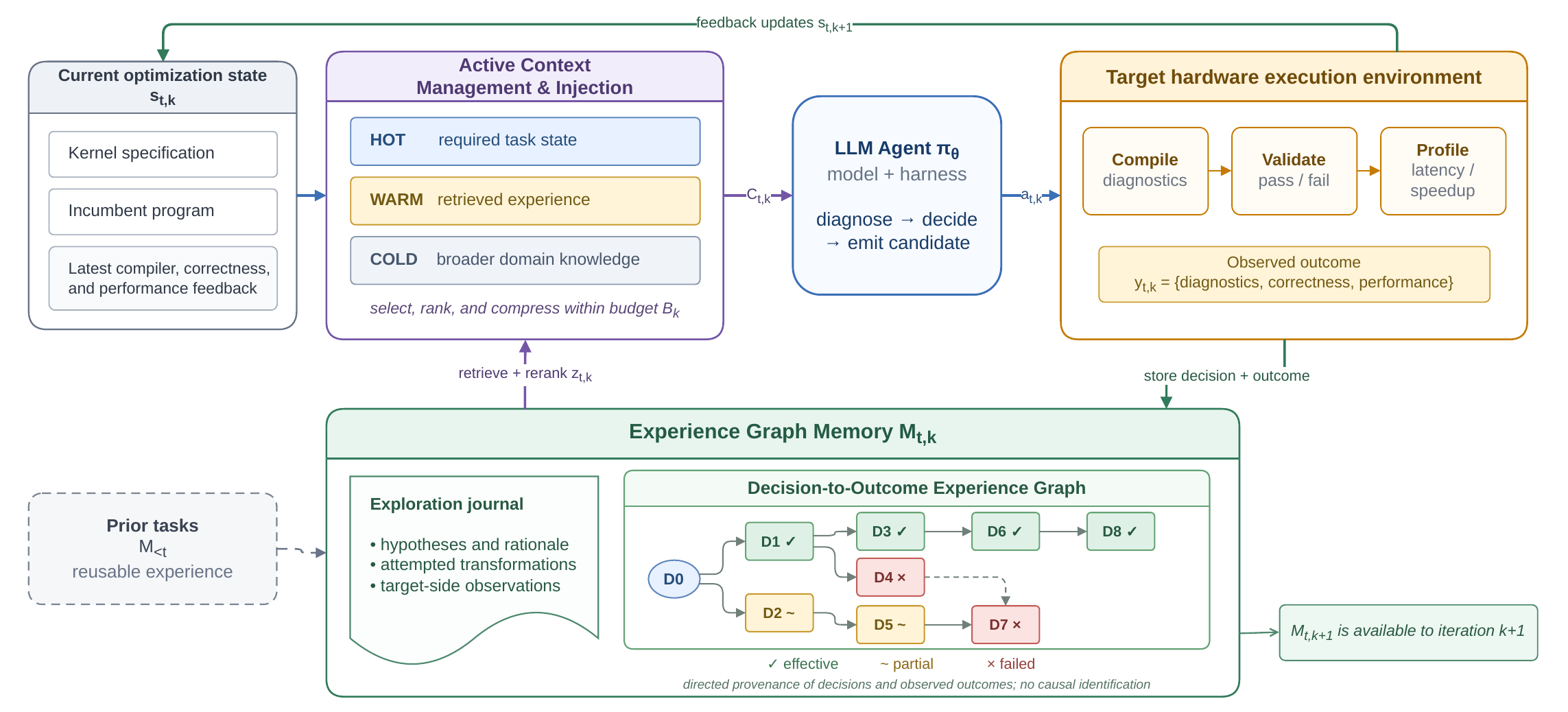}
\caption{KOPE's experience-driven workflow. With model parameters fixed, Active Context Management and Injection constructs a bounded prompt from required task state (hot), retrieved experience (warm), and broader knowledge (cold). Target-side execution returns compiler, correctness, and performance outcomes, which are recorded in a persistent exploration journal and Decision-to-Outcome Experience Graph. Retrieval informs the next decision within the current task and carries experience across tasks.}
\label{fig:kope-loop}
\end{figure*}

Figure~\ref{fig:kope-loop} summarizes the resulting loop. Experience Graph Memory records each decision with its observed outcome and provenance. Active Context Management and Injection then retrieves and supplies a bounded subset of that experience for the current state. These two mechanisms connect target-side execution feedback to subsequent optimization decisions both within and across tasks.

\subsection{Experience Graph Memory}
\label{sec:experience-graph}

Kernel optimization produces branching trajectories. Several attempts may start from the same incumbent, and the value of an earlier decision may become clear only after observing its successors. Experience Graph Memory stores these events with their decision order, observed outcomes, and alternative branches. The links record execution provenance without asserting causal identification.

\paragraph{Dual representations.}
\kopemem{} records each event as both a Journal and a structured Case. The append-only Markdown Journal summarizes the attempted strategy, code change, validation status, measurements, and interpretation. The JSON Case retains the operator and input scene, optimization thought, optional before/after code, compiler or runtime feedback, correctness, absolute speedup, confidence, and session, step, and predecessor identifiers. ReMe provides the general procedural-memory substrate~\cite{cao2026reme}: it distills narrative material and supplies semantic digests, while Cases remain the source of truth for measured outcomes and graph links. Exact Case retrieval remains available if semantic retrieval is unavailable.

\paragraph{Decision-to-Outcome Experience Graph.}
Let $\mathcal{G}=(V,E)$ denote the graph reconstructed from Cases. Each $v\in V$ records a decision and its outcome. An edge $(u,v)\in E$ indicates that the attempt in $v$ began from the optimization state associated with $u$. The current schema stores one \texttt{prev\_case\_id} per Case, so a node has at most one predecessor and may have several successors. The resulting structure is a directed acyclic forest whose depth and branching pattern follow the exploration process. Alternative attempts can share a predecessor. The edge records lineage, not a causal effect.

\paragraph{Graph-aware retrieval and ranking.}
Retrieval first builds a relevance pool. Exact filters match operator, optimization layer, thought identifier, or outcome label. Semantic search supplements this pool with Journals and procedural digests for transfer across tasks. The graph then contributes a separate outcome signal. Let $a_v$ be the absolute speedup of Case $v$ and
\[
r_v=\frac{a_v}{a_{\operatorname{prev}(v)}},
\]
with $r_v=1$ for a root or a nonpositive predecessor value. The schema's \texttt{effect} field stores an observed step-outcome label, not a causal estimate: effective for $r_v\geq1.05$, partial for $1\leq r_v<1.05$, ineffective for $0.85\leq r_v<1$, and negative for $r_v<0.85$. Correctness remains a separate field, and an incorrect Case cannot promote the incumbent.

To account for later outcomes, KOPE uses a discounted downstream outcome score
\[
d(v)=w_{\mathrm{self}}r_v+w_{\mathrm{succ}}\gamma
\sum_{u\in\operatorname{succ}(v)}d(u),
\]
with $w_{\mathrm{self}}=w_{\mathrm{succ}}=0.5$ and $\gamma=0.7$. A terminal Case reduces to its own step ratio. Within each outcome-label bucket, $d(v)c_v$, where $c_v$ is the recorded confidence, orders Cases with otherwise similar relevance. The score is a retrieval heuristic, not an estimate of causal contribution.

\subsection{Active Context Management and Injection}
\label{sec:context-management}

Experience affects a decision only when the relevant records reach the model at the appropriate optimization state. KOPE therefore rebuilds the prompt before every iteration. The query changes with the incumbent, latest diagnostics, and current target. An event recorded at iteration $k$ can thus alter the context supplied at iteration $k+1$.

\paragraph{Dynamic context tiers.}
The three context tiers specify admission priority for the current prompt, not permanent storage locations. \emph{Hot context} contains the kernel specification, incumbent, latest feedback, and optimization target. These items are required whenever the prompt is feasible. \emph{Warm context} contains Cases, Journal summaries, and procedural digests retrieved for the current state, including successful actions and observed failure modes. \emph{Cold context} contains broader documentation, cross-category experience, and older digests. An item may be warm for one state and cold for another.

\paragraph{Budget-aware assembly and injection.}
For a model window of $W$ tokens, system prompt size $S$, generation allowance $G$, and recency reserve $R$, the available knowledge budget is
\[
B_{\mathrm{k}}=\max(0,W-S-G-R).
\]
Let $H$ be the required hot payload, and let $Q_{\mathrm{w}}$ and $Q_{\mathrm{c}}$ be the warm and cold caps. A prompt is feasible only if $H\leq B_{\mathrm{k}}$. The assembler reserves $H$, admits at most $Q_{\mathrm{w}}$ of graph-ranked warm experience, and uses the remaining capacity for at most $Q_{\mathrm{c}}$ of cold material. Selected experience appears as compact Case fields, before-and-after transformations, Journal summaries, or procedural digests. The evaluation fixes these caps for each reported model and context configuration. Learning the caps online lies outside the present implementation.

\paragraph{Boundary to inference-time cache management.}
Active Context Management and Injection selects external material before a model call. KV-cache eviction selects internal key-value states during decoding~\cite{bui2026makeeachcount}. The two mechanisms operate at different stages and can be combined.

% =============================================================================
\section{Experimental Setup}
\label{sec:setup}
% =============================================================================

\subsection{Benchmark and Execution Environment}

We evaluate AscendC kernels with CANN Bench v0.4.0~\cite{gao2026cannbench} on Ascend 910C hardware~\cite{liao2021ascend}. The evaluator release used for these runs contains 53 operators and 20 cases per operator, giving 1{,}060 public cases~\cite{gao2026cannbench}. We use this complete set as the task universe for every configuration. A missing operator contributes 20 failed cases and zero score. Exact-intersection diagnostics distinguish the quality of returned operators from task coverage.

The evaluator checks compilation and numerical correctness before recording performance and applying safeguards against reward hacking, including suspected CPU fallback~\cite{gao2026cannbench}. An operator may retain a positive speedup field even when the evaluator later invalidates its performance score. We recompute all reported values from archived final-evaluation artifacts, excluding intermediate optimization logs.

\subsection{Evaluation Configurations}

We compare KOPE, CANNBot, and CUDA-Agent on the same 53-operator target set under GLM-5.2 and Deepseek-V4-Pro. Each system comparison fixes the model, benchmark, and 1M-token budget. We select these two models to support reproducibility because both provide a 1M-token context window, allowing the context budget to remain fixed across systems. Table~\ref{tab:collected-results} reports the best final artifact for each model and system configuration. Tables~\ref{tab:active-context} and~\ref{tab:graphmem} report the active-context and Experience Graph Memory ablations, respectively. Because the final-evaluator JSON leaves model, agent, context budget, and context-construction policy blank, we use the author-maintained run mapping for these labels.

\paragraph{CANNBot baseline.}
We use CANNBot as the external baseline, with the official CANN samples and optimization repository as its implementation reference~\cite{cannbot2026}. Its GLM-5.2 and Deepseek-V4-Pro runs use the same CANN Bench final evaluator as KOPE. The benchmark remains fixed at 53 operators, so operators absent from a returned artifact count as failures.

\paragraph{Strong GPU-optimization baseline.}
CUDA-Agent reports state-of-the-art KernelBench results on NVIDIA GPUs: a $98.8\%$ pass rate and a $2.60\times$ geometric-mean speedup over PyTorch Eager~\cite{dai2026cudaagent}. It develops kernel-optimization capability from 6{,}000 synthesized tasks and agentic reinforcement learning, and its ReAct-style test loop refines each kernel independently. We evaluate the same general workflow with Deepseek-V4-Pro and GLM-5.2 using Ascend compilation, testing, and profiling. The comparison asks whether model capability and within-operator refinement suffice when target-specific public corpora are sparse. It is not a CUDA-to-Ascend code-transfer test: CUDA-Agent and KOPE are general optimization workflows rather than hardware-specific systems.

\subsection{Ablation Study Settings}

\paragraph{Ablation 1: Active Context Management and Injection.}
We compare two complete GLM-5.2 configurations with different context-construction policies. In the passive condition, the agent must discover relevant knowledge and assemble its own context during optimization. In the active condition, the workflow selects, manages, and injects task-relevant knowledge using the three-tier policy and task-conditioned experience retrieval described in Section~\ref{sec:context-management}. Both archived artifacts contain all 53 operators and 1{,}060 cases, so the comparison uses the complete suite without intersection filtering. We report aggregate optimization token consumption together with correctness and performance. This workflow-level ablation evaluates the integrated context mechanism and does not separately identify the effects of retrieval ranking, context compression, and injection.

\paragraph{Ablation 2: Experience Graph Memory.}
The component ablation fixes the GLM-5.2 workflow and target kernels while disabling or enabling graph-structured memory. Four artifacts, one for each difficulty level, span all 53 operators. They pair the implementation produced without Experience Graph Memory (\texttt{paper\_res\_baseline}) with the graph-enabled implementation (\texttt{cannbench\_ascendc}). The two conditions provide 458 and 431 positive timing measurements, respectively, from the fixed 1{,}060-case suite. A total of 412 cases have positive measurements in both conditions. We report full-suite correctness for both configurations and compute performance only on this exact timing intersection. For paired case $i$, $q_i=T_{\mathrm{without},i}/T_{\mathrm{with},i}$, so a geometric mean above one favors Experience Graph Memory.

\subsection{Metrics and Artifact Provenance}

For the fixed 1{,}060-case benchmark, \emph{pass rate} is $P/1{,}060$, where $P$ is the number of cases accepted by the final evaluator after its reward-hacking checks. Missing operators add no passes and receive zero score. For each returned operator $o$, the artifact reports $s_o=T_{\mathrm{CANN}}/T_{\mathrm{generated}}$, where $T_{\mathrm{CANN}}$ is the latency of the CANN library baseline. The geometric mean excludes missing and nonpositive speedup fields, which can make a small returned subset appear strong. We report it as a descriptive statistic and do not interpret it as suite-wide coverage. \emph{Token use} is the aggregate optimization token consumption recorded across a complete 53-operator configuration. In the CUDA-Agent evaluation, a \emph{reached operator} has at least one passing case. Solving an operator requires all 20 cases. The benchmark's \emph{overall score} combines compilation, correctness, and performance~\cite{gao2026cannbench}, with absent operators assigned zero. Exact-intersection results are secondary because conditioning on returned operators may favor systems with low coverage.

% =============================================================================
\section{Results}
\label{sec:results}
% =============================================================================

\subsection{Collected System Results}

Table~\ref{tab:collected-results} reports the best completed final-evaluation artifact for each of the six model and system configurations. Every pass rate uses the full 1{,}060-case denominator. Cases assigned to missing operators count as failures.

\begin{table}[t]
\centering
\caption{Best full-suite result for each model and system configuration. Op. Pass reports operators with at least one passing case over the 53-operator suite. Case Pass reports accepted cases over all 1{,}060 cases, with missing operators counted as failures.}
\label{tab:collected-results}
\footnotesize
\setlength{\tabcolsep}{2.5pt}
\renewcommand{\arraystretch}{1.08}
\begin{tabular}{@{}llrrrr@{}}
\toprule
\textbf{Model} & \textbf{System} & \shortstack{\textbf{Op.}\\\textbf{Pass}} & \shortstack{\textbf{Case}\\\textbf{Pass}} & \shortstack{\textbf{Pass}\\\textbf{Rate}} & \textbf{Score} \\
\midrule
\multirow{3}{*}{GLM-5.2}
 & \textbf{KOPE} & \textbf{52/53} & \textbf{897/1{,}060} & \textbf{84.6\%} & \textbf{2004.49} \\
\cmidrule(l){2-6}
 & CANNBot & 37/53 & 613/1{,}060 & 57.8\% & 1465.93 \\
\cmidrule(l){2-6}
 & CUDA-Agent & 13/53 & 156/1{,}060 & 14.7\% & 312.00 \\
\midrule
\multirow{3}{*}{\shortstack[l]{Deepseek-\\V4-Pro}}
 & \textbf{KOPE} & \textbf{47/53} & \textbf{783/1{,}060} & \textbf{73.9\%} & \textbf{1242.90} \\
\cmidrule(l){2-6}
 & CANNBot & 25/53 & 417/1{,}060 & 39.3\% & 1015.38 \\
\cmidrule(l){2-6}
 & CUDA-Agent & 8/53 & 96/1{,}060 & 9.1\% & 192.00 \\
\bottomrule
\end{tabular}
\end{table}

\begin{figure*}[t]
\centering
\includegraphics[width=0.98\textwidth]{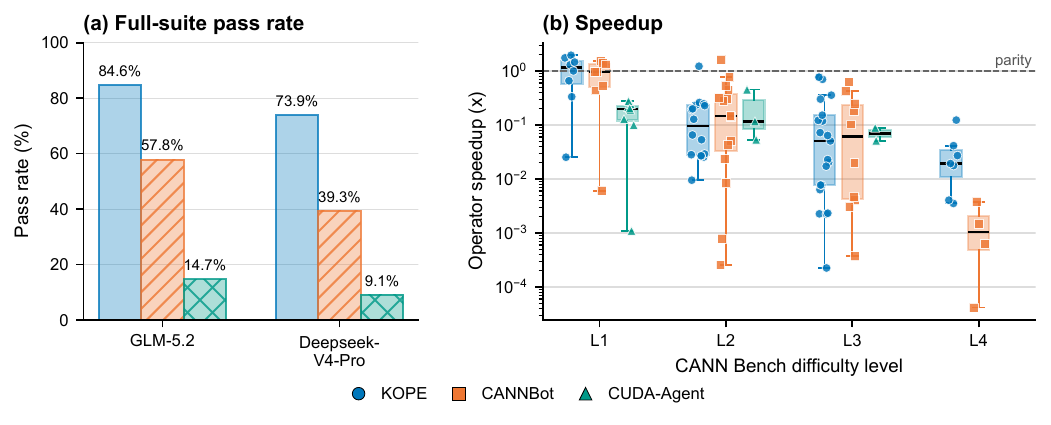}
\caption{Model and workflow comparison on CANN Bench. (a) Full-suite pass rate for KOPE, CANNBot, and CUDA-Agent over 53 operators and 1{,}060 cases. Cases from unreported operators count as failures. (b) Positive GLM-5.2 operator-speedup fields grouped by CANN Bench difficulty and workflow. Boxes show the interquartile range, black lines show the median, whiskers extend to $1.5\times$ the interquartile range, and points are individual measurements. The logarithmic axis marks parity at $1\times$. KOPE, CANNBot, and CUDA-Agent contribute 52, 37, and 13 measurements, respectively. CUDA-Agent contributes 8/3/2/0 measurements from L1 to L4, all below parity. The plot omits missing and nonpositive fields and does not include Deepseek-V4-Pro speedups.}
\label{fig:model-workflow-results}
\end{figure*}

\paragraph{Full-suite GLM-5.2 comparison.}
KOPE returns all 53 operators and passes 897 cases. CANNBot returns 49 operators and passes 613 cases. The artifact reports $62.6\%$ over its 980 returned cases. Once the four missing operators add 80 failures, the full-suite pass rate is $57.8\%$. KOPE improves full-suite pass rate by $26.8$ percentage points and raises overall score from 1465.93 to 2004.49 ($+36.7\%$). On the exact 49-operator intersection, KOPE passes 846 cases ($86.3\%$) and CANNBot passes 613 ($62.6\%$). Their matched scores are 1869.95 and 1465.93, a difference of $27.6\%$. Under the headline operator-speedup aggregation, KOPE's geometric mean is $1.54\times$ CANNBot's.

\paragraph{Full-suite Deepseek-V4-Pro comparison.}
KOPE returns all 53 operators, passes 783 cases, and obtains a $73.9\%$ full-suite pass rate with a score of 1242.90. CANNBot returns 31 operators, reaches 25 of them, and passes 417 of 620 returned cases ($67.3\%$). The 22 missing operators add 440 failures, giving a $39.3\%$ full-suite pass rate and a score of 1015.38. KOPE therefore leads by $34.5$ percentage points in pass rate and $22.4\%$ in score. On the 31-operator intersection, KOPE passes 487 cases and scores 952.53, whereas CANNBot passes 417 and scores 1015.38. CANNBot also has a slightly higher positive-field geometric mean on this intersection (0.0485 over 25 fields versus 0.0434 over 29 fields). CANNBot therefore remains competitive on the returned subset, whereas KOPE provides broader functional coverage and a higher full-suite score.

\subsection{CUDA-Agent in the Sparse-Corpus Ascend Setting}

Both CUDA-Agent runs return all 53 requested operators. With GLM-5.2, CUDA-Agent reaches 13 operators and passes 156 cases ($14.7\%$). With Deepseek-V4-Pro, it reaches 8 and passes 96 ($9.1\%$). Each reached operator passes 12 of 20 cases, and the Deepseek-V4-Pro set is a subset of the GLM-5.2 set. Neither run solves a complete operator. The 13 GLM-5.2 speedup fields comprise 8, 3, and 2 operators at L1, L2, and L3. All are below parity, with a $0.1000\times$ geometric mean. The eight Deepseek-V4-Pro fields (7 at L1 and 1 at L2) all exceed parity and yield $15.5078\times$. These conditional statistics cover different small subsets and exclude 40 and 45 operators, so they do not characterize full-suite performance. Under the same models, KOPE passes 897 and 783 cases. Model choice changes performance on the reached subset but does not produce broad coverage. Together with KOPE's higher pass counts, this pattern is consistent with the value of collecting and reusing target-side evidence in sparse-corpus settings. The comparison does not measure CUDA-to-Ascend knowledge transfer.

\subsection{Ablation Study 1: Active Context Management and Injection}

Table~\ref{tab:active-context} reports the complete-suite active-context comparison. Moving from passive agent-led context construction to active workflow-level context management and injection increases passed cases from 636 to 897 and pass rate from $60.0\%$ to $84.6\%$, a gain of $24.6$ percentage points. The benchmark score rises from 636.00 to 2004.49, and the positive-field geometric mean rises from 0.0382 to 0.0661 ($1.73\times$). At the same time, aggregate optimization token consumption falls from 15.9B to 1.113B, a $93.0\%$ reduction. Both artifacts cover all 53 operators, so the comparison requires neither failure imputation for missing outputs nor intersection selection.

\begin{table}[t]
\centering
\caption{Active versus passive context construction. Speedup reports the geometric mean over positive fields; Tokens reports aggregate use.}
\label{tab:active-context}
\small
\resizebox{\columnwidth}{!}{%
\begin{tabular}{lrrrrr}
\toprule
\textbf{Context Policy} & \textbf{Case Pass} & \textbf{Pass Rate} & \textbf{Score} & \textbf{Speedup} & \textbf{Tokens} \\
\midrule
Passive & 636/1{,}060 & 60.0\% & 636.00 & 0.0382$\times$ & 15.9B \\
Active & 897/1{,}060 & 84.6\% & 2004.49 & 0.0661$\times$ & 1.113B \\
\midrule
Effect & +261 & +24.6 pp & +1368.49 & $1.73\times$ & $-93.0\%$ \\
\bottomrule
\end{tabular}
}%
\end{table}

The active configuration improves accepted cases, score, and conditional speedup while using fewer tokens than the passive configuration. The gain therefore does not come from placing a larger accumulated history into every prompt. Because the active condition jointly changes context selection, compression, and injection, the ablation supports Active Context Management and Injection as an integrated mechanism but does not isolate its internal operations.

\subsection{Ablation Study 2: Experience Graph Memory}

Table~\ref{tab:graphmem} reports per-level geometric means from the positive per-operator speedup fields in the complete GLM-5.2 KOPE artifact: 8, 16, 21, and 7 fields at L1 through L4. We derive Without by dividing With by matched timing ratios computed from 160, 111, 113, and 28 positive-timing pairs.

\FloatBarrier

\begin{table}[H]
\centering
\caption{Experience Graph Memory ablation. Case Pass and Pass Rate use the full 1{,}060-case suite. With reports actual per-level geometric means of positive per-operator speedup fields; Without is derived using matched positive-timing ratios. Effect reports With relative to Without.}
\label{tab:graphmem}
\small
\resizebox{\columnwidth}{!}{%
\begin{tabular}{@{}lrrrrrr@{}}
\toprule
\multicolumn{1}{c}{\multirow{2}{*}[-0.3ex]{\shortstack[c]{\textbf{GraphMemory}\\\textbf{Configuration}}}} & \multicolumn{1}{c}{\multirow{2}{*}[-0.3ex]{\textbf{Case Pass}}} & \multicolumn{1}{c}{\multirow{2}{*}[-0.3ex]{\textbf{Pass Rate}}} & \multicolumn{4}{c}{\textbf{Speedup}} \\
\cmidrule(lr){4-7}
& & & \multicolumn{1}{c}{\textbf{L1}} & \multicolumn{1}{c}{\textbf{L2}} & \multicolumn{1}{c}{\textbf{L3}} & \multicolumn{1}{c}{\textbf{L4}} \\
\midrule
\textbf{With} & \textbf{897/1{,}060} & \textbf{84.6\%} & \textbf{0.659$\times$} & \textbf{0.0866$\times$} & \textbf{0.0344$\times$} & \textbf{0.0182$\times$} \\
Without & 585/1{,}060 & 55.2\% & 0.664$\times$ & 0.0595$\times$ & 0.0229$\times$ & 0.00199$\times$ \\
\midrule
Effect & +312 & +29.4 pp & $0.993\times$ & $1.455\times$ & $1.501\times$ & $9.175\times$ \\
\bottomrule
\end{tabular}
}%
\end{table}

Enabling Experience Graph Memory raises accepted cases from 585 to 897 and full-suite pass rate from $55.2\%$ to $84.6\%$, a gain of $29.4$ percentage points. The configuration without Experience Graph Memory is marginally faster at L1 ($+0.7\%$), but its derived speedup falls below the graph-enabled configuration by $31.3\%$, $33.4\%$, and $89.1\%$ at L2, L3, and L4, respectively. The corresponding matched timing ratios favor Experience Graph Memory by $1.455\times$, $1.501\times$, and $9.175\times$ at these levels; the L4 estimate rests on only 28 paired cases. Across all 412 timing pairs, Experience Graph Memory yields a $1.434\times$ geometric-mean speedup. The correctness result covers the complete suite, whereas the acceleration claim remains conditional on the $38.9\%$ of CANN Bench with positive timings in both configurations.

\FloatBarrier

% =============================================================================
\section{Discussion}
\label{sec:discussion}
% =============================================================================

\subsection{System-Level Comparison}

Across both models, KOPE's main system-level advantage is coverage. Counting missing operators as failures, it leads CANNBot by $26.8$ percentage points in full-suite pass rate with GLM-5.2 and by $34.5$ points with Deepseek-V4-Pro. On the 31-operator Deepseek-V4-Pro intersection, however, CANNBot obtains the higher conditional score and a slightly higher positive-field geometric mean. KOPE therefore improves coverage and correctness across the complete suite without dominating every conditional performance statistic. Because the system comparison changes several workflow decisions at once, it does not attribute the gain to context assembly, trajectory ranking, or another individual component.

CUDA-Agent's conditional speedups vary sharply by model, yet its full-suite pass rate remains between $9.1\%$ and $14.7\%$, and neither configuration solves a complete operator. Conditional speedup on a small reached subset therefore cannot stand in for coverage. The result is consistent with the value of accumulating target-side evidence, but implementation differences prevent attributing the gap to cross-operator memory alone. The comparison also does not imply that either workflow is restricted to one hardware family.

\subsection{Scope of the Evidence}

The active/passive context result comes from one complete pair of configurations. Each model and system cell in Table~\ref{tab:collected-results} is represented by its best archived job, and the Experience Graph Memory result comes from one full-suite correctness comparison and one 412-case paired timing comparison on CANN Bench and Ascend 910C. Repeated paired runs are required to estimate run-to-run variance. Positive-field speedup distributions, particularly for CUDA-Agent, describe reached subsets of different sizes and cannot establish suite-wide acceleration. The available evidence associates active workflow-level context construction with greater complete-suite functional coverage and lower token consumption than passive agent-led construction, and graph-structured memory with higher full-suite pass rate and better aggregate performance on mutually valid paired cases.

% =============================================================================
\section{Conclusion}
\label{sec:conclusion}
% =============================================================================

KOPE turns knowledge acquired during exploration into reusable optimization evidence while keeping the foundation model fixed. Its experience-driven workflow combines Experience Graph Memory, which records ordered decision-to-outcome histories and alternative branches, with Active Context Management and Injection, which supplies relevant experience under a fixed token budget.

The ablations support both mechanisms. Active rather than passive context construction raises full-suite pass rate from $60.0\%$ to $84.6\%$, improves the positive-field geometric mean by $1.73\times$, and reduces optimization token consumption from 15.9B to 1.113B. Experience Graph Memory raises pass rate from $55.2\%$ to $84.6\%$ and yields a $1.434\times$ geometric-mean speedup on 412 paired timing cases. At the system level, KOPE attains a per-operator speedup geometric mean $1.54\times$ that of CANNBot with GLM-5.2 and achieves higher full-suite pass rates with both evaluated models. CUDA-Agent reaches only $9.1\%$ to $14.7\%$ of the Ascend suite and solves no complete operator. Although each configuration is represented by one archived run, the results support KOPE's central claim: for hardware with sparse public training data, retaining and selectively reinjecting target-side experience improves kernel optimization.

% =============================================================================
% Appendix
% =============================================================================
\appendix
\section{Measured Cross-Hardware Retargeting of KOPE on RISC-V}
\label{app:riscv}
% Appendix fragment included by main.tex.

% =============================================================================
\subsection{Scope and Evidence Boundary}
% =============================================================================

The main paper presents KOPE as an experience-driven kernel-optimization workflow rather than an Ascend-specific optimizer. Its primary experiments use AscendC kernels on Ascend NPUs with CANN Bench~\cite{gao2026cannbench}. This appendix evaluates the same workflow in a substantially different stack: Triton kernels~\cite{triton2019} from FlagGems~\cite{flaggems}, executed on a SpacemiT K3 RISC-V processor and compared with PyTorch native kernels~\cite{pytorch2019}. Retargeting changes the hardware (NPU to RVV CPU), kernel language (AscendC to Triton), compiler path, and evaluation harness at the same time.

\paragraph{Evidence boundary.}
Unlike the earlier scenario analysis, the four GLM-5.2 artifacts used here contain completed case-level K3 outcomes for cold start, Ascend-source memory, K3-target memory, and their combination. We recompute every reported value from the archived \texttt{passed}, \texttt{latency\_kope}, and \texttt{speedup\_kope} fields. The export schema retains a legacy metadata object named \texttt{calibration}; none of its target descriptors is read by the analysis or used in an aggregate. The evidence is nevertheless limited to one archived result per setting. The files store one latency value per case rather than the underlying timing samples, and they do not retain numerical-error traces.

% =============================================================================
\subsection{Experimental Design}
% =============================================================================

\begin{figure*}[t]
\centering
\includegraphics[width=0.98\textwidth]{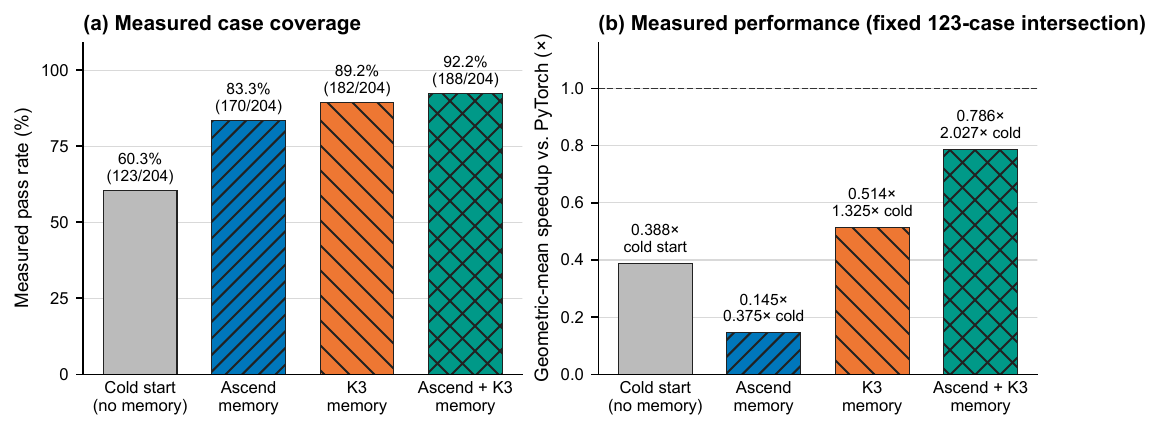}
\caption{Measured outcomes for four KOPE knowledge settings on K3. Panel (a) reports pass coverage over the common 204-case universe. Panel (b) reports geometric-mean speedup over the fixed 123-case intersection passed by every setting, preventing coverage differences from changing the performance denominator. The dashed line marks PyTorch parity. Error bars are omitted because the archive contains per-case point estimates but not the underlying repeated timing samples.}
\label{fig:kope-summary}
\end{figure*}

\subsubsection{RISC-V target environment}

The K3 integrates eight SpacemiT X100 RISC-V cores with the RVV~1.0 vector extension~\cite{riscvv,spacemitk3}. Table~\ref{tab:env} records the pinned software environment.

\begin{table}[t]
\centering
\caption{RISC-V experimental environment.}
\label{tab:env}
\footnotesize
\setlength{\tabcolsep}{2pt}
\begin{tabular}{@{}ll@{}}
\toprule
\textbf{Item} & \textbf{Value} \\
\midrule
Hardware & SpacemiT K3 (8$\times$ X100, riscv64, RVV 1.0) \\
Memory & 16 GB \\
System & Bianbu 4.0.4 (resolute) \\
Python & 3.12.13 \\
PyTorch & 2.8.0+spacemit.0 \\
Triton & 3.6.0+spacemit.a5 \\
FlagGems & pinned release 5.3.0 \\
\bottomrule
\end{tabular}
\end{table}

\subsubsection{Benchmark and aggregation}

The evaluation uses the kernel-only core suite from the FlagGems benchmark and \texttt{triton.testing.do\_bench} for timing. It covers fp32 and fp16; bf16 is disabled by the K3 backend. K3-specific shapes cap each tensor at 16.7M elements because the upstream core shape set targets much larger GPU memories. The common universe contains 204 cases from 21 operators. For configuration $c$ and case $i$,
\[
S_{c,i} =
\frac{T_{\mathrm{PyTorch},i}}
     {T_{\mathrm{KOPE},c,i}},
\]
so $S_{c,i}>1$ favors the KOPE-produced kernel. A case contributes to pass coverage when its archived \texttt{passed} value is true; every such case has a positive latency and speedup in the four files.

Performance aggregation requires care because the settings pass different numbers of cases. We therefore report (1) a conditional geometric mean over all passed cases in each setting and (2) a fixed-intersection geometric mean over the 123 cases passed by all four settings. The latter holds the case set constant when comparing settings. Table~\ref{tab:kope-summary} reports both views.

\subsubsection{KOPE knowledge settings}

\begin{figure*}[t]
\centering
\includegraphics[width=0.94\textwidth]{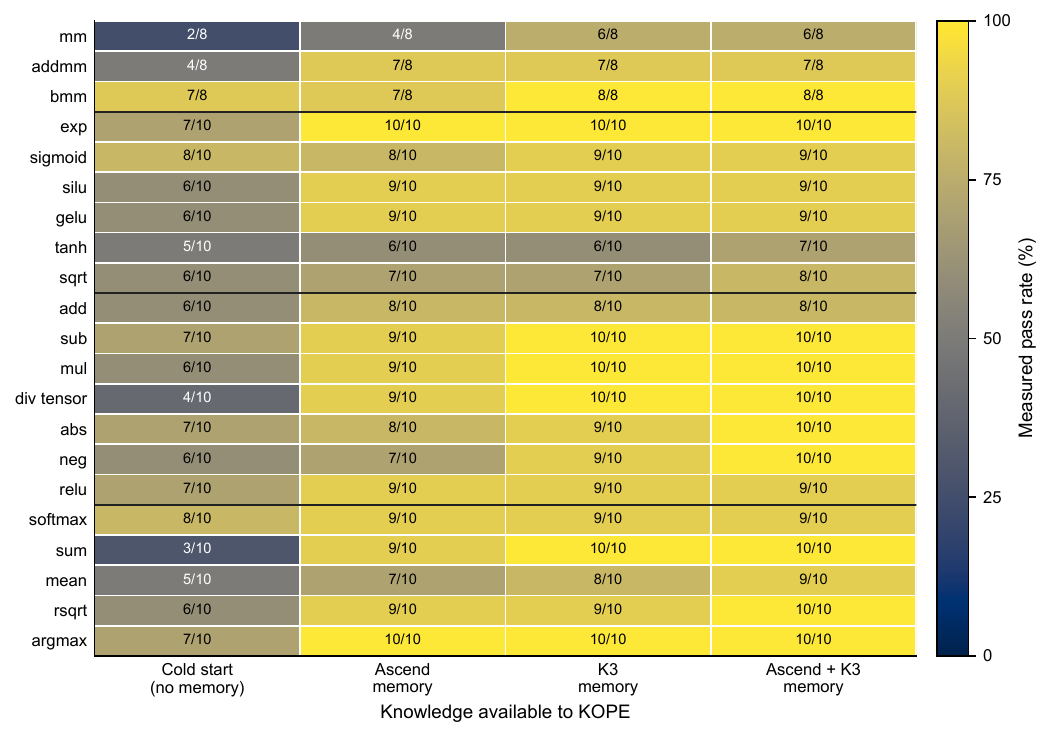}
\caption{Measured passed cases by operator and KOPE knowledge setting. Each cell shows passed/total cases; color encodes the corresponding pass rate. Rows are grouped as matmul, activation, elementwise, and reduction families. The archived pass sets are nested across the four settings; this is an observed property of these runs, not an assumption used by the analysis.}
\label{fig:kope-coverage}
\end{figure*}

All settings use GLM-5.2 and the same case definitions. \textbf{Cold start} provides no prior experience. \textbf{Ascend memory} supplies experience collected on the source stack and directly tests whether source-hardware knowledge remains useful after the hardware, kernel language, and compiler path change. \textbf{K3 memory} represents a second target run with experience learned during earlier K3 exploration. \textbf{Ascend + K3 memory} makes both sources available. This design separates source-memory reuse from target-side adaptation. Figures~\ref{fig:kope-summary} and~\ref{fig:kope-coverage} are generated directly from the four JSON files.

\begin{table}[t]
\centering
\caption{Measured KOPE outcomes on the K3 case universe.}
\label{tab:kope-summary}
\footnotesize
\setlength{\tabcolsep}{2.4pt}
\begin{tabular}{@{}lrrrr@{}}
\toprule
\textbf{Setting} & \textbf{Passed} & \textbf{Rate} & \textbf{GM$_\mathrm{pass}$} & \textbf{GM$_\cap$} \\
\midrule
Cold start & 123/204 & 60.3\% & 0.388$\times$ & 0.388$\times$ \\
Ascend memory & 170/204 & 83.3\% & 0.142$\times$ & 0.145$\times$ \\
K3 memory & 182/204 & 89.2\% & 0.508$\times$ & 0.514$\times$ \\
Ascend + K3 & 188/204 & 92.2\% & 0.758$\times$ & 0.786$\times$ \\
\bottomrule
\end{tabular}
\end{table}

% =============================================================================
\subsection{Retargeting Results}
% =============================================================================

\subsubsection{K3 presents a nontrivial optimization target}

The measured FlagGems baseline reaches or exceeds PyTorch parity in 76 of 204 cases ($37.3\%$), with an overall geometric-mean speedup of $1.004\times$. The near-parity aggregate hides sharply different regimes. Reliable fp32 geometric means for \texttt{bmm}, \texttt{mm}, and \texttt{addmm} range from $30.9\times$ to $48.7\times$, whereas several activations provide smaller gains of $1.08$--$1.65\times$. Most simple elementwise and reduction kernels remain below parity, spanning approximately $0.19$--$0.78\times$. \texttt{layer\_norm} also exposes a Triton compiler defect, and \texttt{rms\_norm} fails during benchmark-shape parsing.

This mixture matters for KOPE. The target is executable, but its optimization space is neither uniformly solved nor dominated by one failure mode. Compilation defects, large positive transformations, small-shape launch costs, memory-bandwidth losses, and reduction-path failures provide distinct positive and negative outcomes for experience memory. The RISC-V stack therefore exercises KOPE's build--measure--record loop rather than serving as a trivial code-portability example.

\subsubsection{Source memory transfers coverage, not speed}

Figure~\ref{fig:kope-summary} compares the four knowledge settings at the suite level.

Cold start passes 123 cases ($60.3\%$). Supplying only Ascend-derived memory passes 170 ($83.3\%$), adding 47 cases and 23.0 percentage points. All 123 cold-start passes remain in the Ascend-memory pass set. This is direct cross-hardware evidence that source experience can broaden the set of accepted solutions on a target with a different ISA and programming stack.

The performance result is different. On the fixed intersection, Ascend memory obtains a $0.145\times$ geometric mean, compared with $0.388\times$ at cold start, a $62.5\%$ decrease. The all-passed conditional means show the same ordering ($0.142\times$ versus $0.388\times$). Ascend knowledge therefore transfers feasibility information more reliably than performance choices. The result supports workflow and knowledge-interface generality, but it rejects a stronger claim that source-hardware experience is automatically performance-portable.

\subsubsection{Target experience recovers performance}

K3-local memory passes 182 cases ($89.2\%$) and reaches $0.514\times$ on the fixed intersection, $1.325\times$ the cold-start value. Combining Ascend and K3 memory reaches 188 passes ($92.2\%$) and $0.786\times$, adding six passes over K3 memory and improving the fixed-intersection geometric mean by $1.530\times$. The combined setting is $2.027\times$ cold start on the same 123 cases. Cases at or above PyTorch parity among each setting's passed cases increase from 17 at cold start to 19, 37, and 60 across Ascend, K3, and combined memory. The combined result remains below parity in aggregate, so it indicates substantial target-side adaptation rather than a solved backend.

\subsubsection{Coverage gains span operator families}

Aggregate values could hide a result driven by a few favorable kernels. Figure~\ref{fig:kope-coverage} therefore reports measured pass coverage for every operator.

Ascend memory improves family-level coverage from 54.2\% to 75.0\% for matmul, 63.3\% to 81.7\% for activations, 61.4\% to 84.3\% for elementwise kernels, and 58.0\% to 88.0\% for reductions. Examples include \texttt{mm} (2/8 to 4/8), \texttt{silu} and \texttt{gelu} (6/10 to 9/10), \texttt{div\_tensor} (4/10 to 9/10), and \texttt{sum} (3/10 to 9/10). The breadth matters more to a general-purpose workflow claim than a gain concentrated in matmul alone.

K3 memory reaches full coverage for \texttt{bmm}, \texttt{exp}, \texttt{sub}, \texttt{mul}, \texttt{div\_tensor}, \texttt{sum}, and \texttt{argmax}. The combined setting additionally completes \texttt{abs}, \texttt{neg}, and \texttt{rsqrt}, while improving \texttt{tanh}, \texttt{sqrt}, and \texttt{mean}. This pattern is consistent with source experience supplying broadly reusable feasibility cues and target experience resolving backend-specific choices.

% =============================================================================
\subsection{Implications for KOPE's Generality}
% =============================================================================

\paragraph{Interface portability is observed, not inferred.}
KOPE's core loop consumes a candidate kernel and receives compilation, acceptance, and latency feedback. Experience Graph Memory records the decision, outcome, and provenance without requiring an AscendC-specific node type. The K3 runs instantiate this contract through Triton, FlagGems, and PyTorch. KOPE can therefore execute its optimization and memory loop in a second, heterogeneous toolchain.

\paragraph{Cross-hardware reuse has two distinct outcomes.}
The Ascend-memory comparison isolates source-to-target reuse. Its 23.0-point coverage gain shows that prior experience remains operationally useful after the hardware and programming model change. Its fixed-case performance loss shows that feasibility knowledge and performance knowledge do not transfer equally. Treating these outcomes separately avoids turning interface portability into an unsupported speedup claim.

\paragraph{Target-side learning is the stronger performance result.}
K3 memory improves both coverage and fixed-case performance over cold start, and the combined setting improves both again. These measurements match KOPE's intended role as a continual optimizer: portable memory provides an initial prior, while target-side feedback revises that prior for the new backend. The supported generality claim is therefore that KOPE is a retargetable, general-purpose optimization workflow that can reuse and refine experience across toolchains. The results do not establish hardware-independent speedup.

% =============================================================================
\subsection{Conclusion}
% =============================================================================

The measured RISC-V study strengthens the main paper's generality claim while narrowing its meaning. KOPE executes its optimization and memory loop on a second stack with different hardware, kernel language, compiler path, and evaluator. Ascend memory raises pass coverage from 60.3\% to 83.3\%, demonstrating useful cross-hardware reuse, but reduces fixed-case performance. K3 memory and combined memory then raise both coverage and performance, reaching 92.2\% coverage and a $0.786\times$ fixed-intersection geometric mean. The evidence supports KOPE as a retargetable, general-purpose workflow for reusing and adapting optimization experience. It does not support a hardware-independent performance guarantee.

% =============================================================================
% References
% =============================================================================
\bibliography{kope_aaai27,support_rv}

\end{document}